\documentclass[journal,comsoc]{IEEEtran}

\usepackage{hyperref}
\usepackage[T1]{fontenc}% optional T1 font encoding
\usepackage{graphicx}
\usepackage{xspace}
\usepackage{algorithm}
\usepackage{algpseudocode}

\usepackage{xcolor}
\usepackage{subfigure}
\usepackage{times}
\usepackage{epsfig}
\usepackage{amsmath}
\usepackage{amssymb}
\usepackage{multirow}
\usepackage{makecell}
\usepackage{arydshln}

\graphicspath{{./figures/}}
\newcommand\Tstrut{\rule{0pt}{2.0ex}}         % "top" strut
\makeatletter
\DeclareRobustCommand\onedot{\futurelet\@let@token\@onedot}
\def\@onedot{\ifx\@let@token.\else.\null\fi\xspace}

\makeatother

\makeatletter
\newcommand*\bigcdot{\mathpalette\bigcdot@{.5}}
\newcommand*\bigcdot@[2]{\mathbin{\vcenter{\hbox{\scalebox{#2}{$\m@th#1\bullet$}}}}}
\makeatother

\ifCLASSINFOpdf
\else
\fi

\usepackage{amsmath}
\usepackage[cmintegrals]{newtxmath}

\begin{document}

\title{ResMatch: Residual Attention Learning for Local Feature Matching}

\author{Yuxin Deng and Jiayi Ma% <-this % stops a space
\IEEEcompsocitemizethanks{\IEEEcompsocthanksitem The authors are with the Electronic Information School, Wuhan University, Wuhan, 430072, China (e-mail: dyx\_acuo@whu.edu.cn, jyma2010@gmail.com).}
}

\markboth{}%
{Shell \MakeLowercase{\textit{et al.}}: ResMatch: Residual Attention Learning for Local Feature Matching}

\maketitle

\begin{abstract}
	Attention-based graph neural networks have made great progress in feature matching learning. However, insight of how attention mechanism works for feature matching is lacked in the literature. In this paper, we rethink cross- and self-attention from the viewpoint of traditional feature matching and filtering. In order to facilitate the learning of matching and filtering, we inject the similarity of descriptors and relative positions into cross- and self-attention score, respectively. In this way, the attention can focus on learning residual matching and filtering functions with reference to the basic functions of measuring visual and spatial correlation. Moreover, we mine intra- and inter-neighbors according to the similarity of descriptors and relative positions. Then sparse attention for each point can be performed only within its neighborhoods to acquire higher computation efficiency. Feature matching networks equipped with our full and sparse residual attention learning strategies are termed ResMatch and sResMatch respectively. Extensive experiments, including feature matching, pose estimation and visual localization, confirm the superiority of our networks. Our code is available at~\url{https://github.com/ACuOoOoO/ResMatch}.
\end{abstract}

%%%%%%%%% BODY TEXT
\section{Introduction}
Establishing point correspondences among images, is a fundamental step in geometric computer vision tasks, such as Simultaneous Localization and Mapping (SLAM)~\cite{fu2021fast} and Structure-from-Motion (SfM)~\cite{schonberger2016structure}, and Multiview Stereo (MVS)~\cite{seitz2006comparison}. The most naive approach to match features, which consist of descriptors, positions and other characteristics~\cite{lowe2004distinctive,mishchuk2017working,detone2018superpoint,dusmanu2019d2,tyszkiewicz2020disk,yan2022learning,barroso2022key}, is searching nearest neighbor (NN) according to the similarity of descriptors. However, NN always suffers from large viewpoint and appearance changes, lack of texture and other problems~\cite{ma2021image}.

Graph-based methods~\cite{caetano2009learning,torresani2008feature,sarlin2020superglue,chen2021learning,shi2022clustergnn,cai2023htmatch,lu2023paraformer} match features with more properties,~\emph{e.g.}, keypoint position, so more correspondences can be found. However, it is notoriously hard to design a graph-matching model that can be easily optimized, while harmoniously bridging multiple information. Benefiting from the tremendous potentials of Deep Learning, SuperGlue~\cite{sarlin2020superglue} elegantly leverages spatial relationship and visual appearance in a Transformer-based graph neural network (GNN)~\cite{vaswani2017attention,dosovitskiy2020image,petar2018graph}. Briefly, SuperGlue fuses descriptor and keypoint position in hyperspace, then uses self- (intra-image) and cross- (inter-image) attention to aggregate the valid information, finally the features augmented by the aggregated information can be precisely matched. Although SuperGlue has achieved inspiring performance and led the current trend in feature matching~\cite{chen2021learning,shi2022clustergnn,cai2023htmatch,lu2023paraformer,sun2021loftr}, it is still unclear how features, which entangle the spatial and visual information, interact during elegant self- and cross-attention propagation.

\begin{figure}
	\centering
	\includegraphics[width=0.95\linewidth]{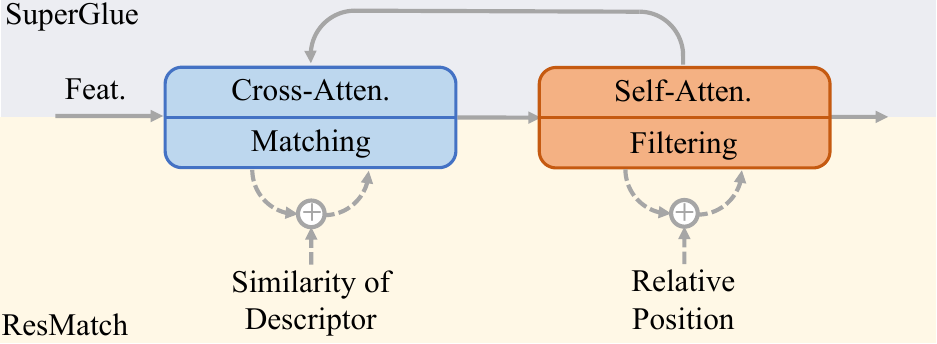}
	\caption{\textbf{Interpretation of ResMatch}, where solid/dash arrow represents feature/attention propagation. From the viewpoint of classical matching-and-filtering, the similarity of descriptors and relative positions are injected into cross- and self-attention.}
	\label{fig:fig1}
\end{figure}

In fact, SuperGlue can be seen as an iterative matching-and-filtering process as shown in Figure~\ref{fig:fig1}. Firstly, it is easy to find cross-attention propagation, which measures the similarity of input features and gathers the message of similar features, behaving like NN. So cross-attention can be deemed as a matching step. Putative sets built by such brute-force matching are too noisy, so match filtering methods~\cite{lowe2004distinctive,bian2017gms,yi2018learning,ma2014robust,lin2017code,yi2018learning,zhang2019oanet,liu2021learnable,ma2014robust} are proposed to remove heavy outliers. VFC~\cite{ma2014robust}, a filter, assumes the vector field consensus,~\emph{i.e.}, optical flow between two images can be represented by a function that can be fitted by only inliers. The function given by VFC has a formulation of a linear combination of kernels, which is similar to attention operation. Moreover, the kernels share the same implication of measuring intra-image correlation with self-attention. So self-attention can be interpreted as a field-consensus-based filtering step. Generally, there is strong evidence for us to promote agnostic attention-based networks from a traditional viewpoint of matching and filtering.

In the iterative attention propagation, cross-attention should learn some residual matching functions in addition to the natural basic function of measuring the similarity of visual descriptors, for example utilizing the estimated field to recall mismatched correspondences. Self-attention might need evaluate the reliability of a correspondence in the image according to the intra-distinction of descriptors, while estimating vector field with spatial information in nonlinear space. From this perspective, cross- and self-attention can be optimized with the respective bypassing prior injection of the similarity of descriptors and the relative positions due to two obvious reasons: ~\emph{Firstly}, correlation or similarity pre-computing can make the attention concentrate on learning residual functions of matching and filtering. ~\emph{Secondly}, spatial and visual information are compressed and mixed up in a limited dimensional space during attention propagation. Conversely, spatial and visual relationships measured at the beginning and connected to attention blocks must be reliable to indicate the matching.

In this paper, we propose ResMatch, in which self- and cross-attention are reformulated as learning residual functions with reference to relative position and descriptor similarity as shown in Figure~\ref{fig:fig1}. Such residual attention learning is totally distinguished from the one for features~\cite{he2016deep}, which might muddy the agnostic features. And it is believed to give vanilla attention-based feature matching networks a formidable and clean inductive bias,~\emph{i.e.}, a prior for easier optimization. Moreover, we propose sResMatch equipped with KNN-based sparse attention to conserve computation cost. Specifically, we search $k$ nearest inter-neighbors for each point as matching candidates according to the similarity of descriptors; $k$ intra-neighbors for partial consensus modeling according to the relative positions. Sparse attention for a point is only computed within its neighborhood. So the computation cost of $\mathcal{O}(N^2)$ in full attention is reduced to $\mathcal{O}(kN)$ for two sets of $N$ points matching. Extensive experiments demonstrate that residual attention in ResMatch can yield significant improvements. And the competitive performance of sResMatch supports our interpretation of attention while reducing the computation cost. Our contributions can be summarized as:

\begin{itemize}
	\setlength{\parskip}{0pt}
	\setlength\itemsep{0.7em}
	\item[-] We propose residual attention learning for feature matching, termed ResMatch. Simple bypassing injection of relative positions and the similarity of descriptors facilitates feature matching learning in practice.
	\item[-] We propose sResMatch, a novel linear attention paradigm based on $k$ nearest neighbor searching, which not only reduces the computation cost but also verifies our analysis of attention for feature matching.
	\item[-] Our ResMatch and sResMatch achieve remarkable performance in feature matching, pose estimation and visual localization tasks.
\end{itemize}

\section{Related Works}
\textbf{Classical feature matching.} NN is the simplest way to match features. However, putative sets yielded by NN always contain too many outliers in challenging scenarios. Therefore, many post-processing methods are studied to clean noisy putative sets, such as mutual nearest neighbor (MNN), ratio test (RT)~\cite{lowe2004distinctive}, filters based on local~\cite{bian2017gms,yi2018learning} or global~\cite{ma2014robust,lin2017code} consensus, and sampling-based robust solvers~\cite{fischler1981random,chum2005matching,barath2020magsac}. Especially, VFC~\cite{ma2014robust}, a global-consensus-based filter whose solution shares a similar formulation with self-attention, encourages us to rethink self-attention from a viewpoint of filtering. Compared to those matching-and-filtering methods, graph-based methods~\cite{caetano2009learning,torresani2008feature} seem more promising because more properties of features are utilized, but the difficulty in modeling and optimization has suppressed their popularity.

\begin{figure*}
	\centering
	\includegraphics[width=0.8\linewidth]{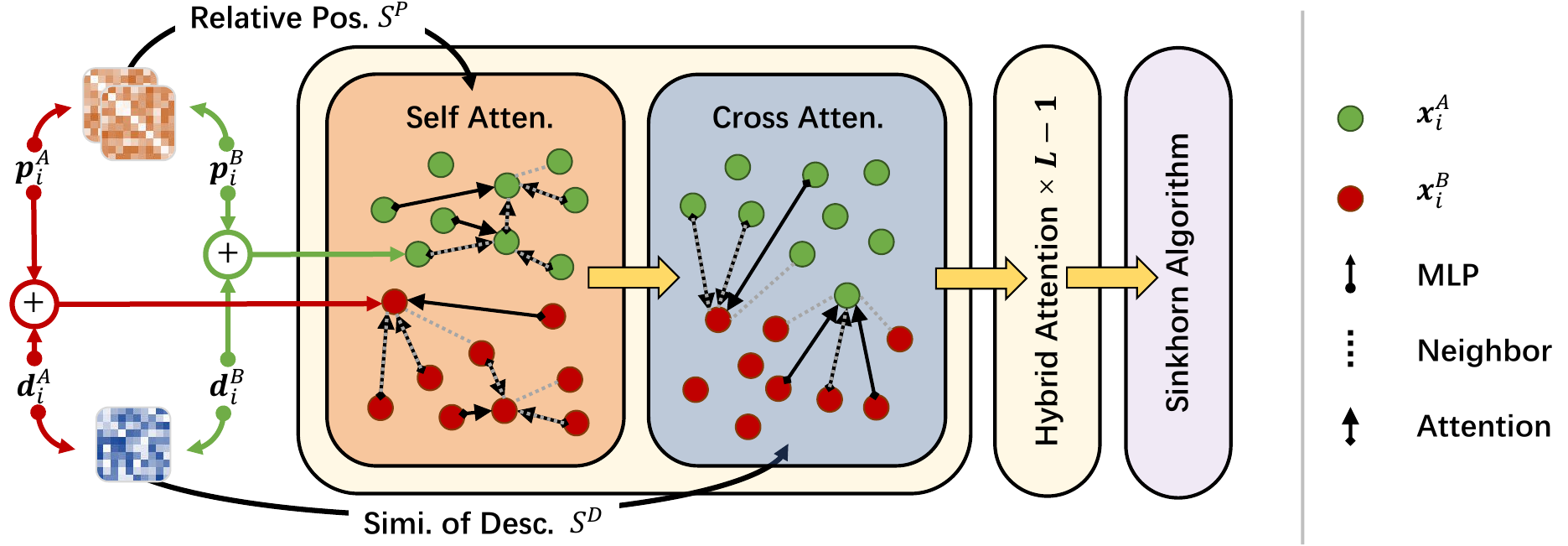}
	\caption{\textbf{The architecture of ResMatch}. Point-to-point relative position and the similarity of descriptors are injected into the self- and cross-attention respectively. The dashed lines link the neighbors mined according to the two bypassing attention, \emph{i.e.}, the relative positions and the similarity of descriptors. Residual attention learned in each layer would correct the bypassing attention and direct the message to pass in the direction of arrows. Note that, the attention would be only computed within neighborhoods in sResMatch.}
	\label{fig:fig2}
\end{figure*}

\textbf{Feature matching learning.} Deep learning has swept many computer vision tasks, including feature matching. PointCN~\cite{yi2018learning} takes an early effort to learn match filtering as a classification task. ConvMatch~\cite{zhang2023convmatch}, an alternative filter, employs self-attention to model vector field consensus~\cite{ma2014robust}, which shares a similar motivation with us. Instead of filtering putative sets, SuperGlue~\cite{sarlin2020superglue} designs an attention-based GNN to match sparse features in a graph matching manner. To improve the practicability of SuperGlue, SGMNet~\cite{chen2021learning} and ClusterGNN~\cite{shi2022clustergnn} simplify $\mathcal{O}(N^2)$ cost of vanilla attention~\cite{vaswani2017attention}, while HTMatch~\cite{cai2023htmatch} and ParaFormer~\cite{lu2023paraformer} study interactive hybrid attentions for better message passing. Unlike these works, we dedicate ourselves to giving insight on attention mechanism. Moreover, instead of extracting sparse features before matching, LoFTR~\cite{sun2021loftr} finds correspondences for dense learnable features and then predicts precise locations for reliable correspondences. LoFTR has caught increasing interest in current years~\cite{chen2022aspanformer,giang2022topicfm,tang2022quadtree,xie2023deepmatcher}, but we believe dependent feature matching still deserves study since it is an essential step in LoFTR's pipeline.

\textbf{Residual learning.} Residual learning, introduced by the well-known ResNet~\cite{he2016deep}, has become a dispensable technique to avoid signal vanishing and facilitate the optimization in deep neural networks. However, most residual connections, including those in feature matching networks, are conducted on features, and only a few works have tried to learn residual attention in the bloom of Transformer~\cite{vaswani2017attention,dosovitskiy2020image}. Realformer~\cite{he2020realformer} is the first to present residual attention learning for Transformer in the natural language processing task, in which attention from previous layers is added into the current layer. EATransformer~\cite{wang2023convolution} enhances the residual attention for a Transformer-based image classification network with learnable convolution. Compared with them, our residual attention is customized according to the nature of matching, and eliminates the intermediate aggregation or enhancement for attention maps.

\textbf{Efficient attention.} Vanilla attention for $N$ point matching takes a computation cost of $\mathcal{O}(N^2)$, which is expensive for real-time applications~\cite{vaswani2017attention,tay2022efficient}. Linear Transformer~\cite{katharopoulos2020transformers} reduces the computation complexity to $\mathcal{O}(N)$ by decomposing the softmax kernel, which is compatible with all transformer-based networks but might degenerate the feature matching performance in practice~\cite{sun2021loftr,viniavskyi2022openglue}. SGMNet~\cite{chen2021learning} searches several putative matches as seeds for down and up attentional pooling, which also achieves linear complexity and obtains satisfying performance. Similarly, the variant of ParaFormer~\cite{lu2023paraformer} uses a part of points for attentional pooling in U-Net architecture~\cite{ronneberger2015u,gao2022graph}. ClusterGNN~\cite{shi2022clustergnn} implements attention within $k_c$ different clusters, which leads to $\mathcal{O}(4N^2/k_c)$ cost at least. Different from other methods, our sResMatch performs sparse attention for each point within a pre-defined neighborhood, which results in linear complexity.

\section{Method}
\subsection{Attention-based Feature Matching Revisit}
Given an image, feature extraction algorithms can yield a set of local features, which consist of $N$ keypoint positions ${\boldsymbol{p}_i \in \mathbb{R}^2}$ and corresponding visual descriptors ${\boldsymbol{d}_i \in \mathbb{R}^c}$, where $i\in\{1,2,\dots,N\}$ and $c$ is the channel of descriptor. Aiming to match two feature sets from image $A$ and $B$, SuperGlue~\cite{sarlin2020superglue} and its alternatives first fuse spatial and visual information in $c$-dimensional space as:
\begin{equation}
	^{0}\boldsymbol{x}_i = f_1(\boldsymbol{d}_i)+f_2(\boldsymbol{p}_i),
	\label{eqn:eqn1}
\end{equation}
where $f(\cdot)$ is a multilayer perceptron (MLP). And then, two sets of fused features $^{0}\boldsymbol{X}^A$ and $^{0}\boldsymbol{X}^B$ are fed into GNN, in which the basic operation, attention feed-forward $\text{Atten}(\boldsymbol{X},\boldsymbol{Y})$ can be described as~\cite{vaswani2017attention}:
\begin{align}
	\label{eqn:eqn2}
	\boldsymbol{Q},\boldsymbol{K},\boldsymbol{V}  & =W_Q(\boldsymbol{X}),W_K(\boldsymbol{Y}),W_V(\boldsymbol{Y}),                  \\
	\label{eqn:eqn3}
	S(\boldsymbol{X},\boldsymbol{Y}) & = \boldsymbol{Q}\boldsymbol{K}^T,                                              \\
	\label{eqn:eqn4}
	\tilde{\boldsymbol{X}}                        & = Softmax(S(\boldsymbol{X},\boldsymbol{Y}))\boldsymbol{V},                              \\
	\label{eqn:eqn5}
	\text{Atten}(\boldsymbol{X},\boldsymbol{Y})   & = \boldsymbol{X}+f_3(\boldsymbol{X}\|W_{\tilde{x}}(\tilde{\boldsymbol{X}})),
\end{align}
where $W(\cdot)$ denotes learnable linear layer and $\|$ denotes concatenation of feature channel.

The attention operation is so-called self-attention if $\boldsymbol{X}$ and $\boldsymbol{Y}$ come from the same image and cross-attention otherwise. In cross-attention, Equations~\eqref{eqn:eqn2} and \eqref{eqn:eqn3} measure the similarity of inter-image features to find matching candidates. Equations~\eqref{eqn:eqn4} and \eqref{eqn:eqn5} gather their message, which behaves as NN matching. If these equations in self-attention only involve the spatial information, the formulation and implication are very similar to the field consensus model introduced by VFC~\cite{ma2014robust}. These evidences encourage us to rethink cross- and self-attention from the viewpoint of a traditional matching-and-filtering pipeline.

After iterative matching and filtering of $L$ layers hybrid attention, information of $^{0}\boldsymbol{X}^A$ and $^{0}\boldsymbol{X}^B$ is propagated into $^{L}\boldsymbol{X}^A$ and $^{L}\boldsymbol{X}^B$, respectively. Finally, the correlation matrix, $W_L(^{L}\boldsymbol{X}^A)W_L(^{L}\boldsymbol{X}^B)^T$, and a learnable dustbin are fed into Sinkhorn algorithm~\cite{cuturi2013sinkhorn,sarlin2020superglue} or dual softmax function~\cite{sun2021loftr} to acquire an assignment matrix. A cross-entropy loss can be constructed to train the network by increasing/decreasing the matching probability of inlier/outlier in the assignment matrix. The architecture of feature matching is briefly illustrated in Figure~\ref{fig:fig2}.

\subsection{Residual Attention Learning}
\textbf{Motivation.} As we analyzed above, SuperGlue can be interpreted as an attention-based iterative feature matching-and-filtering process. In the process, cross-attention blocks should implement matching functions including measuring the similarity of descriptor, recalling mismatched points with field consensus, and so on. Self-attention blocks should estimate field consensus by exploring the spatial relationship and the visual reliability of intra-points~\cite{ma2014robust}. However, linear layers with matrix multiplication in Equations~\eqref{eqn:eqn2} and \eqref{eqn:eqn3} are not robust to measure correlation, which is the basis of matching and filtering functions, let alone more complicated residual functions. Moreover, information entangled by Equation~\eqref{eqn:eqn1} is noisy and not complete. It is difficult to decode by Equation~\eqref{eqn:eqn2} and utilize in attention, which would bother the iterative process.

%in fused feature so that correct matches can be established. Once there is no sufficient information or the function cannot be learned for matching in the first few blocks, the information would turn noisier and harder to be used later. On the other hand, field consensus modeling require stacked neural networks to learn highly nonlinear mapping. So the spatial information also should be clearly passed in feed-forward flow. And an extra learner to map the position might be preferred.

\textbf{Residual cross-attention.} Since the matching step needs to measure the similarity of visual appearance, we directly measure the similarity of raw descriptors of two images as
\begin{equation}
	S^{D}(\boldsymbol{X},\boldsymbol{Y}) = f_4(\boldsymbol{D}^X)f_4(\boldsymbol{D}^Y)^T,
	\label{eqn:eqn6}
\end{equation}
where $\boldsymbol{D}^X$ denotes the descriptor set corresponding to $\boldsymbol{X}$.
And then, we add $S^{D}$ into Equation~\eqref{eqn:eqn3} after affine modulation and activation:
\begin{align}
	\label{eqn:eqn7}
	S^{D'}(\boldsymbol{X},\boldsymbol{Y}) &= \text{LReLU}(\lambda^DS^{D}(\boldsymbol{X},\boldsymbol{Y})+\beta^D), \\
	S^{ResD}(\boldsymbol{X},\boldsymbol{Y}) &= S(\boldsymbol{X},\boldsymbol{Y})+S^{D'}(\boldsymbol{X},\boldsymbol{Y}),
	\label{eqn:eqn8}
\end{align}
where LReLU denotes leaky rectified linear unit, $\lambda$ and $\beta$ are learnable affine parameters unshared in different layers. Note that, the bypassing $S^{D'}$ is broadcast to multi-head $S(\boldsymbol{X},\boldsymbol{Y})$ with unshared affine parameters.

\begin{figure}
	\centering
	\includegraphics[width=1.0\linewidth]{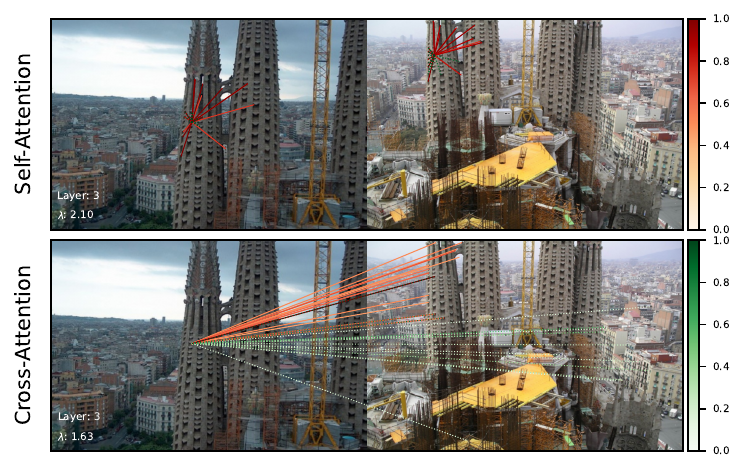}
	\caption{\textbf{Illustration of attention score}. Red full lines denote the top 16 attention scores for the center query point, and green dash lines denote the 16 neighbors according to bypassing attention score. Query points in pictures are correspondences in two views.}
	\label{fig:fig3}
\end{figure}

The visual information in the raw descriptor must be cleaner and more complete than that in the fused feature. And learnable nonlinear mapping $f_4$ is believed more robust for similarity measurement than simple linear mapping in Equation~\eqref{eqn:eqn2}. Therefore, such a bypass makes cross-attention pay less attention to measuring visual similarity and focus on learning residual functions, for example, decoding vector field from features to recall correspondences. As shown in Figure~\ref{fig:fig3}, the similarity of mismatches decreases fast and indicates the cross-attention to some extent.

\textbf{Residual self-attention.} Vector field consensus modeling in self-attention needs clean spatial information and a nonlinear kernel to measure intra-image spatial correlation~\cite{ma2014robust}. So we employ a nonlinear neural network to pre-compute the spatial relationship,~\emph{i.e.}, relative positions as
\begin{equation}
	S^{P}(\boldsymbol{X},\boldsymbol{X}) = f_5(\boldsymbol{P}^X)f_5(\boldsymbol{P}^X)^T,
	\label{eqn:eqn9}
\end{equation}
where $\boldsymbol{P}^X$ denotes the keypoint positions corresponding to $\boldsymbol{X}$. And then, we inject the relative positions into self-attention like residual cross-attention:
\begin{align}
	S^{P'}(\boldsymbol{X},\boldsymbol{X}) &=\text{LReLU}(\lambda^PS^{P}(\boldsymbol{X},\boldsymbol{Y})+\beta^P), \\
	S^{ResP}(\boldsymbol{X},\boldsymbol{X}) &= S(\boldsymbol{X},\boldsymbol{X})+S^{P'}(\boldsymbol{X},\boldsymbol{X}).
	\label{eqn:eqn10}
\end{align}

As shown in Figure~\ref{fig:fig3}, self-attention is assembled in local areas due to the guidance of spatial correlation. Importantly, the topological structures of self-attention of two corresponding points are similar and several correspondences with dark red links can be easily found, which demonstrates that the local consensus is modeled and takes effect during self-attention. More visualization and analysis of attention can be found in Suppl. Material.

\textbf{Bypassing attention adjustment.} As shown in Figure~\ref{fig:fig3}, the similarity of descriptors might not be reliable enough to guide matching in deep layers. And relative positions, which only involve point-to-point spatial relationship but not topological structures in the graphs, might not fulfill the demand of field consensus modeling. Therefore, we adjust these bypassing attentions,~\emph{i.e.}, $S^{D}$ and $S^{P}$, at the middle layer (4th):
\begin{align}
	^4S^{D}(\boldsymbol{X},\boldsymbol{Y}) &= ^0S^{D}(\boldsymbol{X},\boldsymbol{Y})+f_6(^4\boldsymbol{X})f_6(^4\boldsymbol{Y})^T,  \\
	^4S^{P}(\boldsymbol{X},\boldsymbol{X}) &= ^0S^{P}(\boldsymbol{X},\boldsymbol{X})+f_7(^4\boldsymbol{X})f_7(^4\boldsymbol{X})^T,
	\label{eqn:eqn12}
\end{align}
where $f_6$ and $f_7$ are considered as spatial and visual information decoders. $^0S^D$ and $^0S^P$ are computed by Equations~\eqref{eqn:eqn6} and \eqref{eqn:eqn9}, respectively. $^4S^D$ and $^4S^P$ take the places of $^0S^D$ and $^0S^P$ in subsequent residual attention learning.

\subsection{Sparse Residual Attention}
\textbf{Motivation.} Although advanced match filters allow relaxing NN to release more potential correspondences, reconstructing $N^2$ putative matches and gathering all message of them in iterative matching step are too expensive. So it is reasonable to mine only a limited number of candidates according to the similarity of descriptors $S^{D}$ and rematch them in iterations. Moreover, sparse VFC~\cite{ma2014robust} and ConvMatch~\cite{zhang2023convmatch} suggest that global consensus can be represented by only partial bases with higher efficiency and competitive performance. Therefore, if SuperGlue can be seen as an iterative matching-and-filtering process, and bypassing attention dominates in attention propagation, we can search for a sparsification principle based on residual attention learning to acquire a faster solution.

\textbf{KNN-based sparse attention.} We find $k/2$ matching candidates for each point according to $S^{D}$. For filtering, we also mine $k$ nearest neighbors (KNN) for pseudo local consensus verification according to $S^{P}$, instead of selecting partial points as seeds or bases to explore global consensus. So KNN-based sparse attention for a query feature $\boldsymbol{x}_i$ can be formulated briefly as:
\begin{align}
	\label{eqn:eqn14}
	\boldsymbol{M}_i&=\text{KNN}(\boldsymbol{x}_i,S^{*}), S^{*}\in\{S^{P},S^{D}\}, \\
	S(\boldsymbol{x}_i,\boldsymbol{Y}[\boldsymbol{M}_i]) &=	\boldsymbol{Q}\boldsymbol{K}[\boldsymbol{M}_i]^T,
	\label{eqn:eqn15} \\
	\tilde{\boldsymbol{x}_i}            & = Softmax(S/\sqrt{c})\boldsymbol{V}[\boldsymbol{M}_i],
	\label{eqn:eqn16}
\end{align}
where $\boldsymbol{M}$ stores the indices of $k$ nearest neighbor of $\boldsymbol{x}_i$ according to $S^{P}$ or $S^{D}$, and $[\cdot]$ denotes the indexing operation. Note that, the residual attention is not written for clarity, but it is crucial for differential KNN and overall learning. Moreover, KNN is conducted both at the initial step and after bypassing attention adjustment.

Intuitively, more points should be involved in sparse self-attention for global consensus modeling, while the number of matching candidates in cross-attention can be relatively small for distinguishing descriptors. So, at the initial step, we mine $k$ intra- and $k/2$ inter-neighbors for self-attention and cross-attention, respectively; After bypassing attention adjustment, only $k/4$ neighbors are mined for cross-attention. $k$ is empirically set to a constant $64$ for balancing performance and efficiency. In this way, we obtain computation cost linear to the number of points.

\textbf{Comparison to SGMNet}~\cite{chen2021learning}. SGMNet uses NN and RT to acquire seeds for attentional pooling, which results in a linear computation cost as we do. However, we require 6 extra $\mathcal{O}(N^2)$ computations for bypassing attention preparation, versus 2 for SGMNet seeding. To further improve the efficiency, we reduce the number of output channels of $f_{5-7}$ to $c/4$. We believe our method is not sensitive to the reduction of channels because learnable KNN for each point is definitely more robust than handcrafted NN in SGMNet.

\textbf{Comparison to ClusterGNN}~\cite{yan2022learning}. While points are averaged into every cluster, the computation cost of ClusterGNN achieves the low boundary of $\mathcal{O}(4N^2/k_c)$. However, it is doubtful whether fused features would be assembled surrounding the specific centroids of clusters, while the distributions of both keypoint positions and visual descriptors vary apparently among images. Compared to ClusterGNN, our sparse residual attention is more friendly to distribution transfer and has a stable computation cost.

\subsection{Implementation Details}
The network equipped with residual attention learning is termed ResMatch and the corresponding sparse version is termed sResMatch. Our ResMatch and sResMatch consist of sequential $9$ blocks of $4$-head hybrid attention, of which feature dimension is consistent with the input descriptor. Following SGMNet~\cite{chen2021learning}, we train the network on GL3D dataset~\cite{shen2018mirror}, which contains both outdoor and indoor scenes. In the training, 1k features are extracted for each image. 10 iterations of Sinkhorn algorithm are performed to obtain the assignment matrix. And cross-entropy loss is conducted on the final matching probability. The networks are trained in 900000 iterations with batch size of 16. Please refer to Suppl. Material for more details of training.

\begin{figure}
	\centering
	\includegraphics[width=1.0\linewidth]{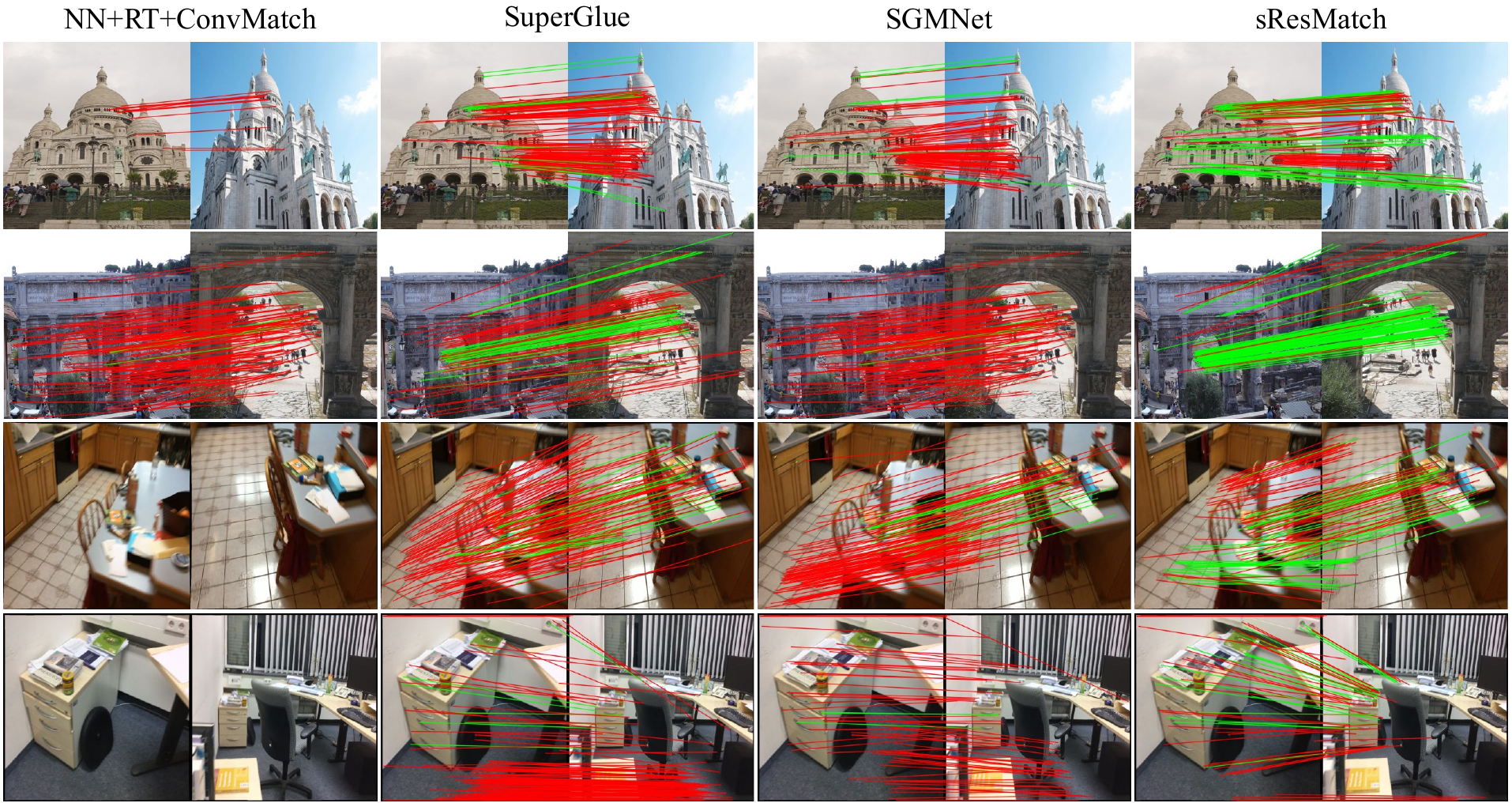}
	\caption{\textbf{Samples of RootSIFT matching results}. 4k features are extracted for images in the second row, and 2k for others. The predicted matches are validated by epipolar constraints. The green/red lines link the inliers/outliers.}
	\label{fig:fig4}
\end{figure}

\section{Experiments}
We evaluate our ResMatch and sResMatch for pose estimation on three datasets, including FM-Bench~\cite{bian2019bench}, YFCC100M~\cite{thomee2016yfcc100m,heinly2015reconstructing}, and ScanNet~\cite{dai2017scannet}. We use Aachen Day-Night~\cite{sattler2018benchmarking} to further verify the applicability of our methods in visual localization task. In all datasets, our methods are employed to match three kinds of features including handcrafted Root-SIFT~\cite{lowe2004distinctive,arandjelovic2012three}, joint DOG+HN\cite{lowe2004distinctive,mishchuk2017working}, and fully learned SuperPoint~\cite{detone2018superpoint}. The performance is compared to classical NN with RT~\cite{lowe2004distinctive}, the state-of-the-art learnable filter ConvMatch~\cite{zhang2023convmatch}, and several feature matching GNNs including SuperGlue~\cite{sarlin2020superglue}, SGMNet~\cite{chen2021learning} and ParaFormer~\cite{lu2023paraformer}. Those feature matching GNNs are trained in the same protocol of our network with $2$-D keypoint position and visual descriptors as input. And a fixed confidence threshold of 0.2 is used to determine matches in the test. Some samples of RootSIFT matching are shown in Figure~\ref{fig:fig4}. Please refer to Suppl. Material for more test details of NN and ConvMatch.

\subsection{Image Matching}
\textbf{YFCC100M} contains 100 million outdoor photos collected from the Internet, along with $72$ 3D models reconstructed in the SfM pipeline~\cite{thomee2016yfcc100m,heinly2015reconstructing}. Following SGMNet~\cite{chen2021learning}, we select 4000 pairs of images from $4$ models for testing. Up to 2k features are extracted for each image by three feature extraction methods and matched by different matchers. Finally, we estimate camera pose with predicted matches and the robust solver, RANSAC~\cite{fischler1981random}. Three metrics are reported in Table~\ref{tab:tab1}: \emph{i)} approximate AUC~\cite{sarlin2020superglue} at different thresholds of estimated rotation and translation error, \emph{ii)} the ratio of the number of correct matches to extracted keypoints, known as matching score, and \emph{iii)} the inlier rate of predicted matches, known as mean precision.

As shown in Table~\ref{tab:tab1}, our residual attention learning brings certain improvements for three kinds of feature matching on most metrics. Especially, we boost matching precision of RootSIFT+SuperGlue and DOG+HN+SuperGlue by over $8\%$ and matching score by about $2\%$, which imposes a positive effect on pose estimation accuracy. It is worth mentioning that KNN-based sResMatch keeps the admirable performance while saving computation costs in theory.

\begin{table}
	\small
	\centering
	\setlength{\tabcolsep}{1.2mm}
	\renewcommand\arraystretch{0.9}
	\caption{\textbf{Results on YFCC100M}, where AUC presents the accuracy of estimated pose, M.S. denotes matching score, and Prec. denotes precision. Bold indicates the best performance.}
	\begin{tabular}{llccccc}
		\Xhline{2\arrayrulewidth} \Tstrut
		\multirow{2}{*}{Feature} & \multirow{2}{*}{Matcher} & \multicolumn{3}{c}{AUC}                               & \multirow{2}{*}{M.S.} &{\multirow{2}{*}{Prec.}}  \\ \cline{3-5} \Tstrut
		&                          & @5$^\circ$  & @10$^\circ$  & @20$^\circ$   &       &       \\ \hline \Tstrut
		\multirow{7}{*}{RootSIFT}
		& NN 											& 48.25 & 58.16 & 68.13 & 4.44  & 56.38 \\
		& ConvMatch                                     & 59.23 & 68.54 & 77.56 & 8.03  & 67.91 \\
		& SuperGlue                                     & 61.45 & 71.23 & 80.62 & 17.37 & 74.91 \\
		& SGMNet                                        & 62.72 & 72.52 & 81.48 & 17.08 & \textbf{86.08} \\
		& ParaFormer                                    & 61.60 & 71.72 & 81.22 & 17.16 & 76.07 \\
		& ResMatch                                      & \textbf{63.92} & 73.33 & 82.15 & \textbf{19.34} & 83.54 \\
		& sResMatch                                     & 63.51 & \textbf{73.52} & \textbf{82.26} & 18.65 & 81.43 \\ \hline \Tstrut
		\multirow{7}{*}{DOG+HN}
		& NN                                        	& 53.05 & 63.11 & 73.37 & 8.44  & 55.96  \\
		& ConvMatch                                     & 59.55 & 69.46 & 75.22 & 13.53 & 60.66     \\
		& SuperGlue                                     & 61.12 & 70.88 & 80.56 & 16.92 & 75.37 \\
		& SGMNet                                        & 62.98 & 72.81 & 82.02 & 17.87 & 83.23 \\
		& ParaFormer                                    & 62.00 & 71.86 & 81.21 & 18.01 & 78.40 \\
		& ResMatch                                      & \textbf{64.48} & \textbf{74.46} & 83.10 & \textbf{19.65} & \textbf{84.91} \\
		& sResMatch                                     & 64.33 & 74.30 & \textbf{83.21} & 18.88 & 81.98 \\ \hline \Tstrut
		\multirow{7}{*}{SuperPoint}
		& NN                                         	& 33.40 & 42.66 & 53.40 & 7.83 & 36.14  \\
		& ConvMatch                                     & 50.28 & 61.25 & 71.78 & 12.58 & 68.34 \\
		& SuperGlue                                     & 60.45 & 70.71 & 80.00 & 19.47 & 80.22 \\
		& SGMNet                                        & 61.22 & 71.02 & 80.45 & 22.36 & \textbf{85.44} \\
		& ParaFormer                                    & 61.75 & 72.03 & 81.23 & 22.31 & 81.28 \\
		& ResMatch                                      & \textbf{62.80} & \textbf{72.81} & \textbf{81.63} & \textbf{23.41} & 84.84 \\
		& sResMatch                                     & 62.45 & 72.22 & 81.44 & 21.99 & 82.61 \\ \Xhline{2\arrayrulewidth}
	\end{tabular}
	\label{tab:tab1}
\end{table}

\textbf{ScanNet} is a comprehensive indoor dataset, composed of monocular sequences with ground-truth poses and depth images~\cite{dai2017scannet}. Unlike those images shot on buildings in YFCC100M, the indoor images in ScanNet test set lack textures and exhibit distinct variance in depth, which obstructs feature matching. Following SuperGlue, we select 1500 wide-baseline pairs of images for the test after overlap validation. For each image, we extract up to 2k RootSIFT and DOG+HN and 1k SuperPoint. The same evaluation pipeline and metrics in YFCC100M are employed in ScanNet.
\begin{table*}[]
	\small
	\centering
	\renewcommand\arraystretch{0.9}
	\caption{\textbf{Results on FM-Bench}, where $\%$Recall reveals the precision of fundamental matrix estimation and $\#$Corr denotes the number of inliers after RANSAC processing.}
	\begin{tabular}{llcccccccc}
		\Xhline{2\arrayrulewidth} \Tstrut
		\multirow{2}{*}{Feature}  & \multirow{2}{*}{Matcher} & \multicolumn{2}{c}{CPC}  & \multicolumn{2}{c}{T\&T}  & \multicolumn{2}{c}{TUM}  & \multicolumn{2}{c}{KITTI}  \\ \cline{3-10} \Tstrut
		&                          & \%Recall    & \#Corr  & \%Recall    & \#Corr   & \%Recall    & \#Corr    & \%Recall     & \#Corr     \\ \hline \Tstrut
		\multirow{7}{*}{RootSIFT}
		& NN                       & 54.6        & 110       & 82.1         & 246       & 62.0        & 454       & 90.3         & 866        \\
		& ConvMatch                & 52.8        & 139         & 73.5            & 228          & 64.4            & 576          & 85.8             & 508           \\
		& SuperGlue                & 58.9        & 239       & 86.8         & 417       & 61.7        & 797       & 90.7         & 1231       \\
		& SGMNet                   & \textbf{62.0}        & \textbf{284}       & 84.6         & 467       & 66.6        & 942       & 90.0         & 1128       \\
		& ParaFormer               & 58.8        & 242       & 86.8         & 427       & 62.4        & 837       & 90.8         & 1263       \\
		& ResMatch                 & 60.7        & 282       & 87.0         & \textbf{474}       & \textbf{67.1}        & \textbf{944}       & 91.1         & \textbf{1318}       \\
		& sResMatch                & 61.7        & 262       & \textbf{87.4}         & 447       & 63.9        & 881       & \textbf{91.2}         & 1292       \\ \hline \Tstrut
		\multirow{7}{*}{DOG+HN}
		& NN                       & 61.9        & 180       & 84.9         & 342       & 59.0        & 658       & 90.6         & 1098       \\
		& ConvMatch                & 55.8        & 181       & 72.7           & 240          & 56.3            &615         & 70.3             &299           \\
		& SuperGlue                & 59.9        & 237       & 87.5         & 417       & 62.9        & 794       & 91.0         & 1219       \\
		& SGMNet                   & 63.0        & 268       & 86.0         & 433       & \textbf{66.6}        & 894       & 90.7         & 1158       \\
		& ParaFormer               & 58.5        & 237       & 87.4         & 414       & 62.3        & 797       & 90.8         & 1204       \\
		& ResMatch                 & \textbf{65.4}        & \textbf{286}       & 87.9         & \textbf{472}       & 65.1        & \textbf{917}       & 91.0         & \textbf{1314}       \\
		& sResMatch                & 65.2        & 266       &\textbf{89.1}         & 454       & 63.1        & 860       & 90.8         & 1303       \\ \hline \Tstrut
		\multirow{7}{*}{SuperPoint}
		& NN                       & 45.8        & 180       & 82.7         & 335       & 54.3        & 563       & 88.0         & 802        \\
		& ConvMatch                & 55.5        & 242       & 85.8         &  394      & 59.9        & 743       & 85.8             &   848         \\
		& SuperGlue                & 70.6        & 354       & 95.0         & 489       & 59.9        & 801       & 88.0         & 971        \\
		& SGMNet                   & 70.7        & \textbf{378}       & 93.5         & \textbf{510}       & \textbf{63.4}        & \textbf{923}       & 87.6         & 984        \\
		& ParaFormer               & 70.4        & 351       & 95.1         & 484       & 60.4        & 806       & 88.5         & 983        \\
		& ResMatch                 & 72.3        & 366       & 95.2         & 503       & 61.4        & 889       & 88.1         & \textbf{1022}       \\
		& sResMatch                & \textbf{74.2}        & 344       & 95.2         & 480       & 57.8        & 739       & \textbf{89.4}         & 974       \\
		\Xhline{2\arrayrulewidth}
	\end{tabular}
	\label{tab:tab3}
\end{table*}

\begin{table}
	\small
	\centering
	\setlength{\tabcolsep}{1.0mm}
	\renewcommand\arraystretch{0.9}
	\caption{\textbf{Wide-baseline indoor pose estimation in ScanNet}. Note that ConvMatch is trained on indoor dataset SUN3D~\cite{xiao2013sun3d}. Bold indicates the best performance.}
	\begin{tabular}{llccccc}
		\Xhline{2\arrayrulewidth} \Tstrut
		\multirow{2}{*}{Feature} & \multirow{2}{*}{Matcher} & \multicolumn{3}{c}{AUC}                               & \multirow{2}{*}{M.S.} &{\multirow{2}{*}{Prec.}}  \\ \cline{3-5} \Tstrut
		&                          & @5$^\circ$  & @10$^\circ$  & @20$^\circ$   &       &       \\ \hline \Tstrut
		\multirow{7}{*}{RootSIFT}
		& NN     										& 21.60 & 30.80 & 40.54 & 2.33  & 28.21 \\
		& ConvMatch                                     & 24.76 & 33.43 & 43.02 & 4.28  & 32.20 \\
		& SuperGlue                                     & 30.08 & 41.30 & 53.00 & 9.08  & 38.44 \\
		& SGMNet                                        & 30.66 & 41.13 & 52.80 & 8.79	& 45.54 \\
		& ParaFormer                                    & 31.26 & 42.40 & 53.85 & 9.10 	& 38.24 \\
		& ResMatch                                      & \textbf{35.40} & \textbf{46.53} & \textbf{57.97} & \textbf{10.19} & 44.94 \\
		& sResMatch                                     & 35.20 & 46.13 & 57.85 & 9.51 	& \textbf{46.09} \\ \hline \Tstrut
		\multirow{7}{*}{DOG+HN}
		& NN                                            & 24.66 & 33.73 & 44.05 & 4.63  & 26.20 \\
		& ConvMatch                                     & 24.70 & 36.13 & 43.12 & 7.77  & 27.85 \\
		& SuperGlue                                     & 34.06 & 44.60 & 56.03 & 8.90 	& 40.36 \\
		& SGMNet                                        & 34.53 & 45.67 & 57.68 & 9.33 	& 45.11 \\
		& ParaFormer                                    & 33.06 & 44.73 & 56.28 & 9.23 	& 40.70 \\
		& ResMatch                                      & \textbf{38.10} & \textbf{49.47} & \textbf{60.80} & \textbf{10.30} & \textbf{45.74} \\
		& sResMatch                                     & 37.13 & 48.47 & 59.61 & 9.72 	& 42.64 \\ \hline \Tstrut
		\multirow{7}{*}{SuperPoint}
		& NN                                            & 22.53 & 30.97 & 41.28 & 12.44 & 31.58  \\
		& ConvMatch                                     & 24.63 & 33.93 & 40.25 & 14.94      & 37.37      \\
		& SuperGlue                                     & 34.18 & 44.35 & 54.89 & 16.25 & 46.16 \\
		& SGMNet                                        & 35.40 & 44.83 & 55.80 & \textbf{16.86} & \textbf{47.83} \\
		& ParaFormer                                    & 34.20 & 44.66 & 55.92 & 16.58 & 46.39 \\
		& ResMatch                                      & 35.80 & \textbf{46.91} & \textbf{58.00} & 16.82 & 47.56 \\
		& sResMatch                                     & \textbf{35.96} & 46.73 & 57.93 & 16.81 & 46.73 \\ \Xhline{2\arrayrulewidth}
	\end{tabular}
	\label{tab:tab2}
\end{table}

Our methods outperform other counterparts with distinct margins for pose estimation as shown in Table~\ref{tab:tab2}. Especially, there is an over $3.5\%$ average improvement at different thresholds for RootSIFT and DOG+HN matching, which demonstrates that the solid inductive bias provided by residual attention learning benefits feature matching in challenging scenes. Especially, the remarkable performance of sResMatch reveals sparsification according to relative positions and the similarity of descriptors is an appropriate and efficient choice for optimizing the self- and cross-attention. And it also confirms our interpretation of the attention-based feature matching network.

\textbf{FM-Bench} comprises four subsets (CPC, T$\&$T, TUM and KITTI) covering driving, indoor SLAM and wide-baseline scenarios~\cite{bian2019bench}. There are 4000 pairs of images in total and we extract up to 4k features for each image with three feature extraction methods. Fundamental matrices are estimated on predicted matches and the estimation is considered correct if its normalized symmetric epipolar distance~\cite{zhang1998determining} is lower than 0.05. In Table~\ref{tab:tab3}, we report recall of fundamental matrix estimation and the mean number of correspondences after RANSAC processing.

As we can see, our methods show advantages on most metrics. However, SGMNet seems to perform better on TUM, while our sResMatch yields a degeneration for SuperPoint matching. That might be caused by low resolution, poor quality and lack of textures in TUM indoor images. Forcing feature extraction methods to extract dense and unreliable features troubles the subsequent matching. Especially, SuperPoint is forced to extract 4k features with a lower threshold of detected score than that in other datasets. In this case, our sResMatch might be struck in local consensus between two unreliable subsets, which finally leads to relatively weak performance.

\subsection{Visual Localization}
Visual localization is a common task to verify the applicability of feature matching algorithms. Thus, we integrate several feature matching algorithms into the state-of-the-art visual localization pipeline Hierarchical Localization (HLoc)~\cite{sarlin2019coarse} and run the pipeline on the Aachen Day-Night V1.0 dataset, which consists of 4328 reference images and 922 (824 daytime, 98 nighttime) query images. For each image, we extract up to 8k RootSIFT, 4k DOG+HN and SuperPoint associated with NetVLAD~\cite{arandjelovic2016netvlad} global feature. To save test time, we only perform 50 iterations of Sinkhorn algorithm for GNN-based methods. The localization accuracy under different thresholds is reported in Table~\ref{tab:tab4}. As we can see, our methods achieve high scores on all metrics, which confirms the applicability of residual attention learning in the downstream tasks. Moreover, our proposals demonstrate scalability to the number of features, as evidenced by the favorable performance on 8k RootSIFT, which is several times larger than the training size of 1k.

\begin{table}[]
	\small
	\centering
	\setlength{\tabcolsep}{1.2mm}
	\renewcommand\arraystretch{0.9}
	\caption{\textbf{Performance in visual localization task}. */*/* denotes the visual localization accuracy under thresholds of {(0.25m, 2$^\circ$)/(0.5m, 5$^\circ$)/(1.0m, 10$^\circ$)}.}
	\begin{tabular}{llcc}
		\Xhline{2\arrayrulewidth} \Tstrut
		\multirow{2}{*}{Feature}  & \multirow{2}{*}{Matcher}   & \multicolumn{2}{c}{(0.25m,2$^\circ$) / (0.5m,5$^\circ$) / (1.0m,10$^\circ$)} \\ \cline{3-4} \Tstrut
		&            & Day               & Night                        \\ \hline \Tstrut
		\multirow{5}{*}{RootSIFT}
		& SuperGlue  & 83.1/ 90.9/ 95.6             & 71.4/ 75.5/ 85.7             \\
		& SGMNet     & 81.7/ 90.5/ 96.1             & 67.3/ 75.5/ 93.9             \\
		& ParaFormer & 83.4/ 91.1/ 95.8             & 69.4/ 75.5/ 87.8             \\
		& ResMatch   & 84.0/ 92.0/ 95.6             & 72.4/ 79.6/ 95.9             \\
		& sResMatch  & 84.8/ 92.6/ 96.6             & 73.5/ 84.7/ 92.9             \\ \hline \Tstrut
		\multirow{5}{*}{DOG+HN}
		& SuperGlue  & 82.8/ 91.0/ 96.1             & 69.4/ 75.6/ 93.9             \\
		& SGMNet     & 84.8/ 92.4/ 97.0             & 74.5/ 84.7/ 98.0             \\
		& ParaFormer & 84.0/ 91.9/ 96.1             & 70.4/ 79.6/ 94.8             \\
		& ResMatch   & 84.1/ 91.7/ 96.8             & 72.4/ 81.6/ 98.0             \\
		& sResMatch  & 85.0/ 93.1/ 97.5             & 75.0/ 85.6/ 98.0             \\ \hline \Tstrut
		\multirow{5}{*}{SuperPoint}
		& SuperGlue  & 86.7/ 93.1/ 97.1             & 77.6/ 88.8/ 98.0             \\
		& SGMNet     & 86.7/ 93.7/ 97.1             & 82.7/ 90.8/ 99.0             \\
		& ParaFormer & 87.7/ 92.4/ 97.2             & 81.6/ 90.8/ 98.0             \\
		& ResMatch   & 87.0/ 93.7/ 97.1             & 82.7/ 92.9/ 98.0             \\
		& sResMatch  & 88.7/ 93.3/ 97.5             & 84.7/ 91.8/ 98.0            \\\Xhline{2\arrayrulewidth}
	\end{tabular}

	\label{tab:tab4}
\end{table}

\begin{table}[]
	\centering
	\setlength{\tabcolsep}{2.0mm}
	\renewcommand\arraystretch{0.9}
	\caption{\textbf{Ablation Study on YFCC100M and ScanNet}. AUC under a threshold of 20$^\circ$ is reported. }
	\begin{tabular}{lcc}
		\Xhline{2\arrayrulewidth} \Tstrut
		AUC@20$^\circ$ ($\%$)     & YFCC100M & ScanNet \\ \hline \Tstrut
		SuperGlue    & 80.62    & 53.00   \\ \hline \Tstrut
		ResMatch     & 82.15    & 57.97   \\
		w/o ResSelfAtten  & 82.00    & 57.16   \\
		w/o ResCrossAtten & 81.35    & 55.47   \\
		w/o Adjustment     & 81.42    & 56.31   \\ \hline \Tstrut
		sResMatch    & 82.26    & 57.85   \\
		w/ FullSelfAtten   & 82.33    & 58.12   \\
		w/ FullCrossAtten  & 81.92    & 58.04   \\
		$k$=32         & 80.78    & 56.34   \\ \Xhline{2\arrayrulewidth}
	\end{tabular}
	\label{tab:tab5}
\end{table}

\section{Discussion}
\subsection{Ablation Study}
In this paper, we propose residual self-, cross-attention and their corresponding sparse versions for feature matching. To investigate the effectiveness of each proposal, we conduct ablation study with 2k RootSIFT on YFCC100M~\cite{thomee2016yfcc100m,heinly2015reconstructing} and ScanNet~\cite{dai2017scannet}. Results of ablation study are reported in Table~\ref{tab:tab5}.

As we can see, ResMatch without residual cross-attention produces larger degeneration of performance than ResMatch without residual self-attention. The reason might be that the information of visual appearance in $c$-D raw descriptors is compressed by $f_1$ to make room for $2$-D positional information in the $c$-D fused hyperspace. The information loss confuses the matching function in ResMatch without residual cross-attention and SuperGlue~\cite{sarlin2020superglue}. Conversely, $2$-D positional information mapped into $c$-D intermediate features has been complete and clean enough for filtering of ResMatch without residual self-attention. However, residual self-attention still improves the SuperGlue with certain margins, which demonstrates the significance of bypassing injection of relative position. Moreover, bypassing attention adjustment would facilitate the learning in deep layers as proved by the ablation study.

For sResMatch, the ablation study suggests that our sparsification principle ($k=64$) of self-attention or cross-attention would not yield distinct changes in performance. We even can find some improvement brought by sparse attention in Tables~\ref{tab:tab3} and \ref{tab:tab4}, in which large numbers of features are extracted. The reason might be that the information of all points is aggregated into a limited feature space. And the mixed information is too ambiguous to model precise vector field. By contrast, the sparsification tightens the solution space of modeling and obtains more precise model after limited iterations. However, too sparse attention with $k=32$ leads to significant drop because matching candidates are too few to cover enough correspondences and the neighborhoods in self-attention are too small to support the global consensus.

\subsection{Computation Efficiency}
\begin{figure}
	\centering
	\includegraphics[width=1.0\linewidth]{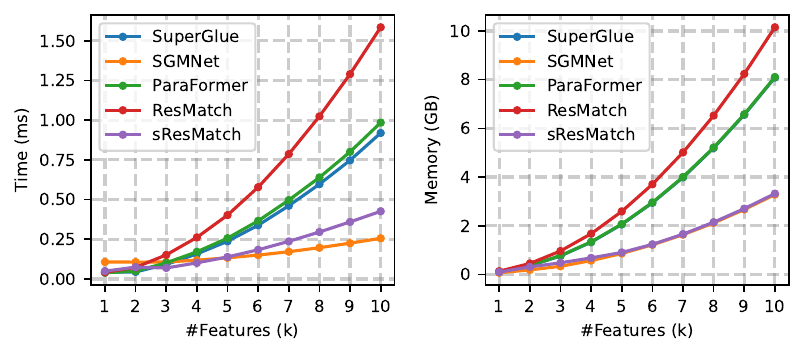}
	\caption{\textbf{Computation efficiency} of feature matching networks on a NVIDIA RTX3090 GPU. The memory consumption of SuperGlue is close to ParaFormer.}
	\label{fig:fig5}
\end{figure}

We compare the computation efficiency of our networks to SuperGlue~\cite{sarlin2020superglue}, SGMNet~\cite{sarlin2020superglue}, ParaFormer~\cite{lu2023paraformer}. The computation cost versus the numbers of 128-D features, are drawn in Figure~\ref{fig:fig5}. Our ResMatch takes the most time and memory to match large numbers of features, despite its significant improvements and simple formulation. That extra time and memory cost relative to SuperGlue is mainly caused by Equations~\eqref{eqn:eqn7} and \eqref{eqn:eqn8}, which require $\mathcal{O}(N^2)$ multiplications. We plan to design cheaper strategies for coupling residual attention in the future. Although the cost of sResMatch is smaller than SGMNet in theory, some operations are hard to optimize in programming. For example, KNN in Equation~\eqref{eqn:eqn14} takes $20\%$ time consumption for 8k features matching and indexing operations in Equations~\eqref{eqn:eqn15} and \eqref{eqn:eqn16} take $23\%$. For memory cost, our sResMatch with $k=64$ is competitive with SGMNet, which is one of the major reasons why $k$ is set to 64.

\section{Conclusion}
In this paper, we rethink self- and cross-attention in feature matching networks from a viewpoint of feature matching and filtering. For the viewpoint, self- and cross-attention are reformulated as learning residual functions with reference to the basic functions of measuring spatial and visual correlation, then added by relative position and the similarity of descriptors to facilitate the learning, respectively. ResMatch, the feature matching network equipped with the proposed residual attention, obtains promising performance in extensive experiments. Furthermore, we conduct sparse self- and cross-attention propagation of each point only with intra- and inter-neighbors, which are mined according to the two kinds of references. sResMatch with sparse residual attention not only reduces the computation cost, but also verifies the significance of residual attention learning with competitive performance.

In summary, we bridge the gap between the interpretable matching-and-filtering pipeline and agnostic attention-based feature matching networks empirically. Comprehensive experiments confirm the validity of our analysis and the superiority of our networks. Although there are still some limitations in our method, we believe this work can advance the development of general feature matching learning.

{\small
	\bibliographystyle{ieee_fullname}
	\bibliography{ResMatch}
}
\end{document}